\documentclass[10pt,twocolumn,letterpaper]{article}

\usepackage{cvpr}              

\usepackage{graphicx}
\usepackage{amsmath}
\usepackage{amssymb}
\usepackage{booktabs}
\usepackage{multirow}

\usepackage[pagebackref,breaklinks,colorlinks]{hyperref}

\usepackage[capitalize]{cleveref}
\crefname{section}{Sec.}{Secs.}
\Crefname{section}{Section}{Sections}
\Crefname{table}{Table}{Tables}
\crefname{table}{Tab.}{Tabs.}

\def\cvprPaperID{13} 
\def\confName{CVPR}
\def\confYear{2022}

\begin{document}

\title{Efficient Multi-Purpose Cross-Attention Based Image Alignment Block for Edge Devices}

\author{Bahri Batuhan Bilecen, Alparslan Fişne, Mustafa Ayazoğlu\\
Aselsan Research\\
Ankara, Turkey\\
{\tt\small \{batuhanb, afisne, mayazoglu\}@aselsan.com.tr}
}
\maketitle

\begin{abstract}
    Image alignment, also known as image registration, is a critical block used in many computer vision problems. One of the key factors in alignment is efficiency, as inefficient aligners can cause significant overhead to the overall problem. In the literature, there are some blocks that appear to do the alignment operation, although most do not focus on efficiency. Therefore, an image alignment block which can both work in time and/or space and can work on edge devices would be beneficial for almost all networks dealing with multiple images. Given its wide usage and importance, we propose an efficient, cross-attention-based, multi-purpose image alignment block (XABA) suitable to work within edge devices. Using cross-attention, we exploit the relationships between features extracted from images. To make cross-attention feasible for real-time image alignment problems and handle large motions, we provide a pyramidal block based cross-attention scheme. This also captures local relationships besides reducing memory requirements and number of operations. Efficient XABA models achieve real-time requirements of running above 20 FPS performance on NVIDIA Jetson Xavier with 30W power consumption compared to other powerful computers. Used as a sub-block in a larger network, XABA also improves multi-image super-resolution network performance in comparison to other alignment methods.

\end{abstract}

\section{Introduction}
\label{sec:intro}

\begin{figure}[h]
    \centering
    \subfloat[Motion image]{\includegraphics[ width=0.49\linewidth]{"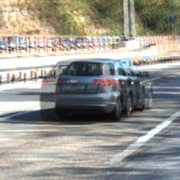"}}
    \hspace{0.01\linewidth}
     \subfloat[DISFlow\cite{Kroeger2016}+warp]{\includegraphics[ width=0.49\linewidth]{"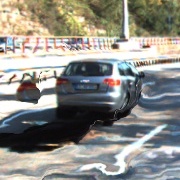"}}\\
     \subfloat[FlowNet\cite{Dosovitskiy2015}+warp]{\includegraphics[ width=0.49\linewidth]{"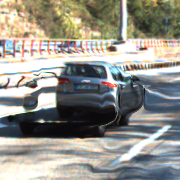"}}
    \hspace{0.01\linewidth}
     \subfloat[XABA]{\includegraphics[ width=0.49\linewidth]{"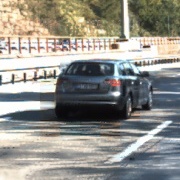"}}
    \caption{XABA's alignment performance compared to DISFlow and FlowNet based alignment. Motion image represents combination of reference and target images.}
    \label{fig:Alignexp2}
\vspace{-1em}
\end{figure}
Image alignment (image registration) aims to align or match images to a chosen reference image. This task constitutes an important part of many computer vision problems dealing with multiple images either in space and/or time, such as restoration~\cite{Yu2020,Wang2019Restoration}, segmentation~\cite{Hu2020}, HDR imaging~\cite{Yan2019}, stereo imaging~\cite{Wang2019}, multi and single-image super-resolution and video super-resolution~\cite{Isobe2020,Xue2019,Wen2022}.

To construct a backbone used in many vision-related problems, there have been many image alignment methods/blocks developed to this day with different focuses. These methods can be divided into three main categories: feature-point extraction-based alignment, classical optical flow-based alignment and deep learning-based alignment.

Feature-point extraction-based alignment algorithms utilize extracted feature points to create a global transform matrix between reference and target frames. Later, the resulting matrix is used to warp frames or images onto each other~\cite{Rublee2011}. Classical optical flow-based alignment algorithms compute the flow vectors between reference frame pixels and target frame pixels, then use the flow vectors in the warping process. These methods can vary greatly, from computing affine transforms of image patches using flow vectors~\cite{Lucas1981} to using a contrast constancy assumption and iteratively reducing misalignments between image pairs~\cite{Brox2004}. Deep learning-based alignment algorithms inherit ideas such as deformable convolutions~\cite{Dai2017} and deep-learning based optical flow~\cite{Dosovitskiy2015}, and attention mechanisms~\cite{Yu2020}.

Despite having been implemented in many ways, most of the aforementioned image alignment methods suffer trading between accuracy and speed, especially on low-end devices. Feature-point extraction-based alignment algorithms require the planar scene assumption, which does not hold most of the time. Although having efficient implementations, classical optical-flow algorithms can fail with large motions. They can also cause bottlenecks when used in deep learning pipelines, as it avoids being able to provide an end-to-end deep learning-based solution. This result may result in sub-optimal solution, since optical flow part is not trained and is based on hand-crafted features. This is indeed the problem of integrating classical algorithms into deep learning pipelines.

On the other hand, having the entire solution blocks in deep learning framework helps using hardware at the hand to the full extent easily by using the already supplied deployment tools such as NVIDIA's TensorRT~\cite{NVIDIA} and Intel's OpenVINO~\cite{OpenVINO}, and improves the performance of the overall trained pipeline. Besides, deep learning-based alignment methods are fairly more recent and have promising advancements. Deformable convolutions add 2D offsets to sampling locations in traditional convolutions, introducing more adaptation and easing the alignment~\cite{Haiyun2019}. Deep learning-based optical flow methods~\cite{Sun2018,Zachary2020} yield accurate alignment; however, they can be quite expensive and hence may not be feasible to run on embedded or edge devices. Recent work on image alignment used attention mechanism because of its native feature matching and transforming properties~\cite{Abati2021,Yu2020}, however direct application of attention operation for image alignment can be memory and computation hungry. The use of attention mechanism started with natural language processing domain with the seminal works ~\cite{Luong2015, Bahdanau2016, Vaswani2017} and later transferred to image domain~\cite{correia2021attention} and it is known to improve the performances of various networks on image domain~\cite{Wang2018} as well. Nonetheless, using attention for image alignment is fairly recent and not much study is present, especially focusing on edge device interference efficiency.

Motivated by the recent performance \& accuracy related studies on attention, the lack of efficient attention-based alignment studies, and the lack of a general-purpose deep learning-based image alignment block applicable to different problems on edge devices; we propose a cross-attention based image alignment block which can be integrated into many deep learning pipelines with ease. We name our proposed method as \textbf{XABA (cross-attention based aligner)}. XABA is designed with efficiency and plug-and-play approach in mind that can run in real-time. We divide reference and target images into non-overlapping sub-images, extract features from each sub-image, and efficiently compute a cross-attention matrix to align sub-images in feature-level. Dividing images into blocks allows us to process them in parallel, ease implementations on embedded systems, and force local information extraction.
In addition, to boost the proposed methods performance in large motions, the baseline block is applied in a pyramidal fashion to the input image at different scales which effectively increases the attended area while keeping the computational cost requirements at minimum. The final output of the network is used by fusing the pyramidal network outputs with a pixel attention based fusion module. We also do not have hand-crafted hyper-parameters for our cost function unlike~\cite{Abati2021}, which further eases the training process.

For more robustness and adaptation to different applications, we also propose a sparsification scheme to calculate the attention matrix, which can be used to find sparse attention matrices. This can be further exploited for fast matrix multiplication during inference. 

Furthermore, as an alternative to the classical softmax non-linearity used in attention matrices, we combined hard-thresholded ReLU (clips above 1) with row-normalization as the activation function. This effectively normalizes the rows of the calculated attention matrix, promotes sparsity, and makes the implementation more suitable to be used in edge devices due to the lightness of the activation function.


The main contributions of this paper are:
\begin{itemize}
    \item We propose a block-based, pyramidal, multi-purpose, deep-learning based image alignment block using cross-attention.
    \item We propose an alternative to softmax activation, combining hard-thresholded-ReLU with Normalization.
    \item The proposed block is efficiency-focused and real-time applicable, proven by multiple tests on edge devices.
\end{itemize}


\section{Related Works}
\label{sec:relatedworks}

{\bf Feature-point based alignment.} This method aims to match feature points extracted from images via ~\cite{Yang2018,Rublee2011,Kuang2015}. Using outlier methods like RANSAC~\cite{Fischler1981} and their derivations ~\cite{Raguram2013,Rahman2019}, some extracted feature points are eliminated. Then, using rest of the matching feature points, a global transformation matrix is generated between images. The major downside of this method is that the motion is restricted to be globally uniform. In other words, every part of the image is being subjected to the same global motion, which is not correct most of the time.

{\bf Traditional \& deep-learning-based optical flow alignment.} Optical flow calculates flow vectors for each pixel between reference image pixels and target image pixels. Since a one-to-one correspondence is obtained between pixels (for occlusion-free regions), images can be aligned accordingly. Optical flow is an under-constrained problem, hence additional constraints like brightness consistency and confining to small movements are required. Lucas-Kanade~\cite{Lucas1981} and Horn-Schunck~\cite{Horn1981} are two of the most well-known classical flow algorithms, with other variations also present~\cite{Sanchez2013}. Classical flow algorithms are still being developed, one of the recent ones being DISFlow~\cite{Kroeger2016} which focuses on time complexity. Deep-learning based optical-flow, on the contrary, are much more recent which gained momentum with FlowNet~\cite{Dosovitskiy2015}. FlowNet proposed using a convolutional network in flow estimation for the first time. FlowNet2~\cite{Ilg2017} offered using correlation layers as an improvement. It also included a new stacked architecture including image warping, and sub-networks for small displacements. PWC-Net~\cite{Sun2018} refined flow in a coarse-to-fine manner and utilized feature warping to reduce the network size. In addition to performing better than FlowNet2, PWC-Net also has a smaller network size. Recent algorithms like RAFT~\cite{Zachary2020} started to utilize correlation volumes and recurrent structures to further increase the flow quality. RIFE~\cite{Huang2020} directly estimates intermediate flow estimations from a low-framerate video for frame interpolation purposes. 
Even though most deep-learning based flow methods have high accuracies, they are not applicable on embedded environments due to their high computational demand at practically meaningful image sizes.

{\bf Attention and attention-based alignment.} Bahdanau~\cite{Bahdanau2016} and Luong~\cite{Luong2015} models, also known as additive and multiplicative/dot-product attentions respectively, are the first remarkable studies about attention. Later on, Google proposed their attention-based network architecture, the Transformer~\cite{Vaswani2017}. All these studies were done on natural language processing; however, attention-based solutions in computer vision problems were also starting to emerge. Inspired by non-local means denoising algorithm~\cite{Buades2005}, Wang et al.~\cite{Wang2018} proposed a non-local building block which captures long-range dependencies within feature maps. This study prepared the ground for many vision applications, such as classification~\cite{Ramachandran2019,Zhao2020}, object detection~\cite{Carion2020,Guo2019}, image segmentation~\cite{Hu2020,Oh2022}, image super-resolution~\cite{Wang2019}, and video super-resolution~\cite{Xue2019,Isobe2020,Wen2022}. Regardless, the idea of attention is fairly new on image alignment algorithms. Only a handful of studies ~\cite{Yu2020,Wen2022,Abati2021} are present, most of which do not focus on efficiency and deployment on embedded environments. 

\section{Proposed Method}
\begin{figure*}
    \centering
    \includegraphics[width=\linewidth]{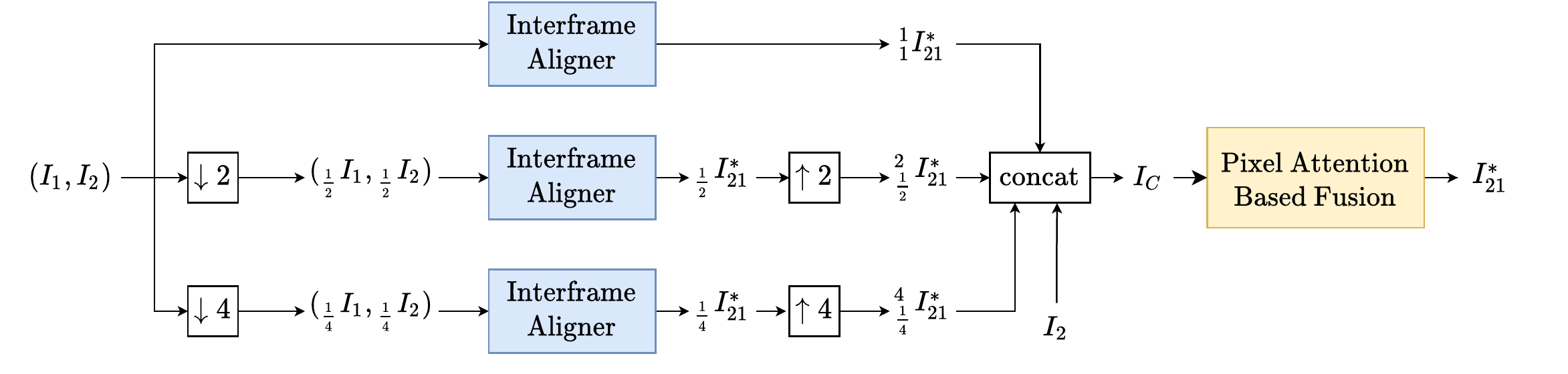}
  \caption{Pyramidal global alignment block. concat denotes tensor concatenation. $\uparrow$ and $\downarrow$ denote upscaling and downscaling operations, respectively.}
  \label{fig:global}
\end{figure*}

\subsection{Background information}
{\bf Attention mechanism.} The attention mechanism proposed in~\cite{Vaswani2017} and~\cite{Luong2015} has three main concepts: key, query and value. 

Query $(Q)$ is the input vector for which the attention is desired to be calculated. Query is compared with all keys $(K)$ by taking the dot products between Q and K, creating a key-query matrix $({QK^T})$. Higher values in the resulting matrix indicate higher correlation between the relevant elements of Q and K.
The matrix is then normalized with some constant $(\sqrt{d_k})$ and its softmax is calculated to make the sum of each row 1. At the end, multiplying with values $(V)$ gives the attention (\ref{equ:att}).

\begin{equation}
    Att(Q,K,V) = \text{softmax}(\frac{QK^T}{\sqrt{d_k}})V = AV
    \label{equ:att}
\end{equation}

In self-attention, $Q$, $K$ and $V$ are equal to each other. Naturally, the equality does not hold for cross-attention. In our study, $Q$ represents the feature matrix of the reference image, the image to be aligned. $K$ is the feature matrix of the target image. $V$ is the reference image itself.

\subsection{Baseline Block - Interframe Aligner}
\label{sec:baselineblock}

\begin{figure}
    \begin{center}
    \includegraphics[width=\linewidth]{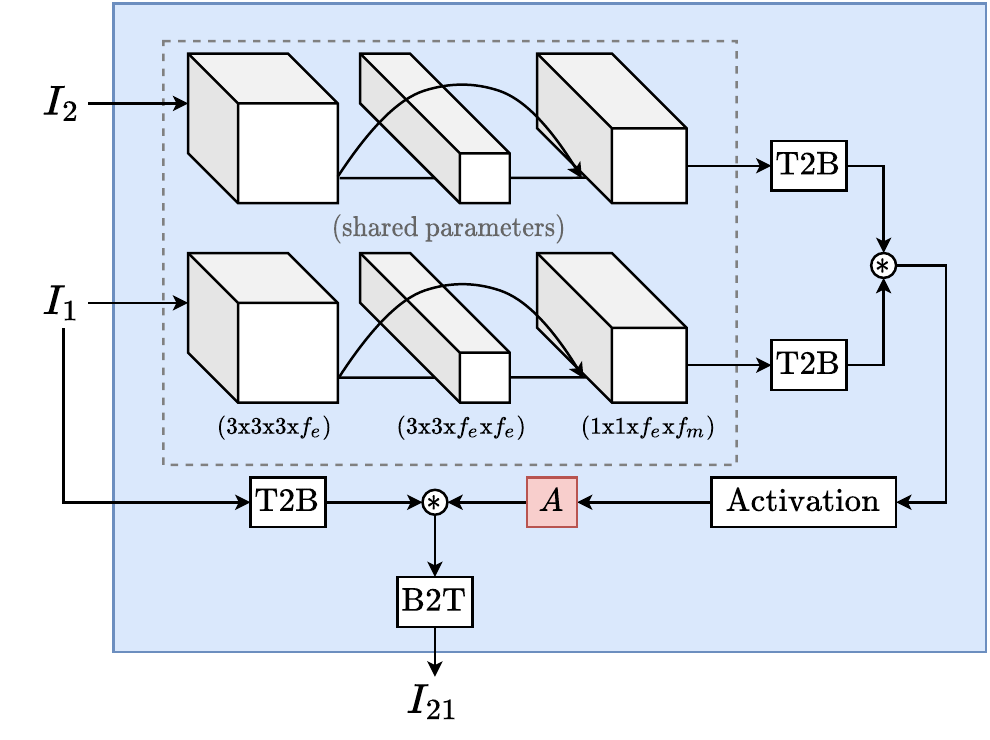}
    \caption{Baseline block - interframe aligner. Reference ($I_{1}$) and target frames ($I_{2}$) are sent as inputs and aligned frame $I_{21}$ is obtained at the output. $f_e$ and $f_m$ are modifiable dimension parameters for feature extraction. T2B is the operator transforming tensors into non-overlapping blocks and reorganize the tensor in the batch dimension, while B2T is the inverse operator. $*$ is matrix multiplication operation. $Activation$ is the function to normalize the attention matrix, $A$.}
    \label{fig:alignment}
    \vspace{-2em}
    \end{center}
\end{figure}

Our \emph{baseline block} structure which we refer as \emph{interframe aligner block} is given in~\cref{fig:alignment}. Using cascaded convolutional layers with skip connections and ReLU activations, features of the input images $(I_{1},I_{2})$ are extracted. Note that the parameters of this residual feature extraction network is shared between images. These features are then convolved with 1x1 kernels to reduce the number of operations. Note that depending on the application, the layer with 1x1 kernel may or may not share the same parameters with its parallel branch. For instance, 1x1 convolutions can be shared in aligning two RGB images as done in our experiments, however; to match images in different domains (such as thermal and RGB images) further adaptations may be needed.

After the feature extraction, resulting features are sent to tensor-to-block (T2B) operator. T2B operator divides the feature images into non-overlapping patches and stacks these blocks in batch dimension. In this sense, it is somewhat similar to pixel unshuffling~\cite{Shi2016} where the unshuffled pixels are stacked on the channel dimension. This enables us to only match spatially closer features with each other while allowing parallel processing. 

To find the correlation between two image features, T2B outputs are matrix multiplied to create a dot-product attention matrix as described in~\cref{equ:att}. After normalization along the rows of the result with a non-linearity (softmax, or hard-thresholded ReLU with Normalization), $A$, the \emph{attention matrix} between two input image features is generated. Note that this matrix $A$ transfers images from K (key) domain to Q (query) domain via an adaptive linear combination and blending. This linear combination map is then used for aligning $I_1$ and $I_2$. Block-to-tensor (B2T) operation is applied at the end to reverse the effects of T2B operator, which is similar to pixel shuffling~\cite{Shi2016} where it operates along the channel dimension.

\begin{figure}[h]
    \begin{center}
    \includegraphics[width=\linewidth]{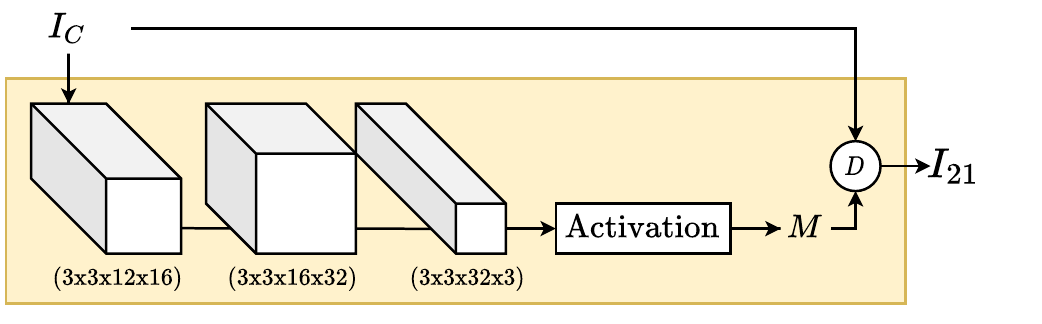}
    \caption{Pixel attention based fusion block. Using the outputs of interframe aligners, it generates a mask $M$ to be used in fusing the aligned images. $D$ denotes the following operation: $D(A,B) = \sum_{i}^{}A[i] \cdot B[i]$, where $A[i]$ and $B[i]$ have the same dimensions for all $i$, and $\cdot$ denotes element-wise matrix multiplication operation. For the resulting tensor after $concat$ operation ($I_C$), each $i^{th}$ dimension represents an aligned image  (${}^{K^{-1}}_KI^{*}_{21}$).}
    \label{fig:masknet}
    \end{center}
    \vspace{-1em}
\end{figure}

As discussed above, the baseline block is good enough to capture and align small displacements between features. However, to be able to capture large motions, non-overlapping block size should be increased which is a parameter of T2B operator. Unfortunately, increasing the block size increases the size of $A$, and hence it increases the computational load and memory requirements which is not suitable for edge devices. As an alternative to handle large motions, we propose a pyramidal processing scheme which is more efficient and suitable for edge devices.

\subsection{Pyramidal Global Alignment Block}
As described in~\cref{sec:baselineblock}, our baseline interframe aligner is suitable for capturing local feature matches. In other words, it is good at capturing small displacement of features between images. To effectively handle large motions and effectively increase the block size while being computationally light, we propose another block which we refer as \emph{Pyramidal Global Alignment Block}. This block encapsulates different number of baseline blocks dedicated to work with different scales of the input images (\cref{fig:global}).

An input image pair $(I_{1},I_{2})$ is sent to the alignment block. Each baseline interframe aligner takes down scaled input image pairs $({}_KI_{1},{}_KI_{2})$, where the downscaling factors are denoted by $K=\{1,\frac{1}{2},\frac{1}{4},...\frac{1}{n}\}$. Interframe aligners generate the aligned frames (${}_K{I^{*}_{21}}$), all of which have different resolutions due to different downscalings. For all outputs of interframe aligners, upsampling is applied with the same scaling factor and therefore are scaled back to their original resolution (${}^{K^{-1}}_KI^{*}_{21}$). The individual outputs of the baseline blocks for different scales are fused into a single image using the Pixel Attention Based Fusion Block. Final aligned frame result, $I^{*}_{21}$ is obtained at the end.

\subsection{Pixel Attention Based Fusion Block}

Individual outputs of the baseline blocks for different scales constitute candidate images and hence candidate pixels. Inherently, at unit scale level, the image resolution is high but only small displacements are handled. At $\frac{1}{2}$ scale, level medium displacements are handled but the image resolution is lower. At $\frac{1}{4}$ scale level, very large displacements are handled; however, the resolution is at its lowest. Given these, a selection mechanism is needed. \emph{Pixel Attention Based Fusion Block} is used for this purpose in such a way that it tries to combine different outputs to form a single image. 

The fusion block inherits a cascaded convolutional network which generates a mask $M$ using all interframe aligner outputs, shown in~\cref{fig:masknet}. 
Concatenated aligned images ($I_C$) are passed through the CNN and the activation function, which performs normalization in the channel dimension. This ensures that the contribution of all interframe aligner output energies are unchanged. Each channel of the resulting mask $M$ are then used as a multiplicative mask for the corresponding images in $I_C$ and all masked images are summed up to obtain the final image. Note that the mask in this case determines the combination ratio of all interframe aligner outputs from different scales. 




\section{Experiments}

For the experiments, we used Kitti~\cite{Menze2015CVPR} dataset which is commonly used in image alignment, stereo and optical flow benchmarking. Kitti includes 200 training and 200 test stereo scene pairs, captured in rural and city traffic. Images are in RGB and lossless png format, with resolutions not the same among all images but all around 1250x375.

The experiments can be divided into three different sections. In the first experiment, we used XABA by itself for image alignment. In the second experiment, we used the image pairs of the Kitti dataset and posed a Multi Image Super-Resolution problem (MISR) and showed the performance of XABA in combination with a Single Image Super-Resolution (SISR) network to solve MISR problem. In the third experiment, for the different parameter settings and block sizes of XABA, we have taken measurements from NVIDIA Jetson Xavier and showed real-time capabilities of the proposed method.

\subsection{Training Details}

For the first and second experiments, we used Adam optimizer with $\beta_1 = 0.9$ and $\beta_2 = 0.999$ and used maximum learning rate = 2e-3 with Knee learning rate scheduling~\cite{Nikhil2020} and warm up for all of our experiments. Mini batch size of 8 is used and the models are trained for 170 epochs, where each epoch consumes the training images 10 times. Each mini-batch is composed of image patches cropped from random images from the training set and standard geometric transformations such as rotate \& flip were used as a form of data augmentation. We used 320x320 as crop size and for the second experiment (\cref{sec:sr_perf}) the low resolution cropped images were created by 4 times downscaling the original images. In both experiments, Charbonnier loss was used with \( \epsilon = 0.1\) as defined in (\ref{mat:charbon}). Charbonnier loss is the smoother version of L1-loss, which is known to have better convergence characteristics than L2-loss.
\vspace{-0.3em}
\begin{gather}
    Charbonnier (x) = \sqrt{x^2+\epsilon^2}
\label{mat:charbon}
\end{gather}






\subsection{Alignment Performance}
\label{sec:alignment_perf}

\begin{figure*}
\begin{center}

\subfloat[Original Reference Image]{\includegraphics[clip, trim=100 50 50 100,width=1\linewidth]{"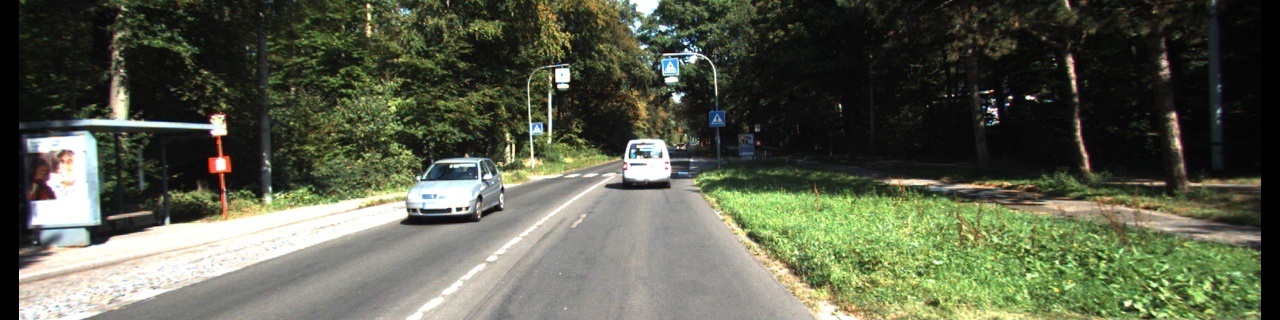"}} \\
\subfloat[Original Target Image]{\includegraphics[clip, trim=100 50 50 100,width=1\linewidth]{"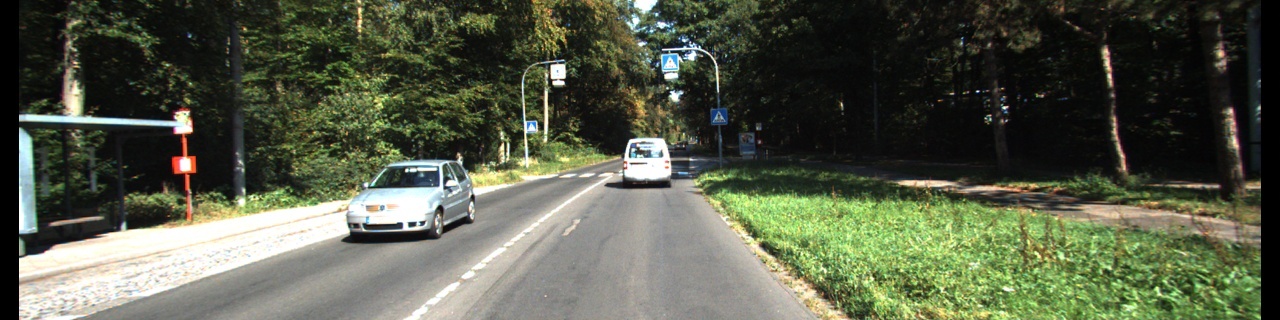"}} \\
\subfloat[Motion Image]{\includegraphics[clip, trim=20 20 20 20,width=0.25\linewidth]{"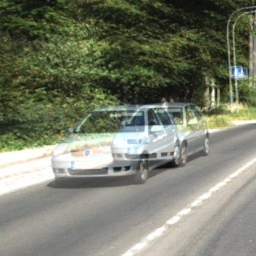"}}
\subfloat[DISFlow]{\includegraphics[clip, trim=20 20 20 20,width=0.25\linewidth]{"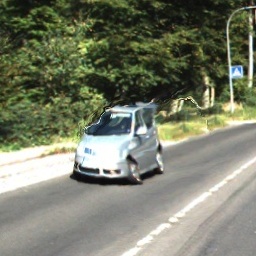"}}
\subfloat[FlowNet]{\includegraphics[clip, trim=20 20 20 20,width=0.25\linewidth]{"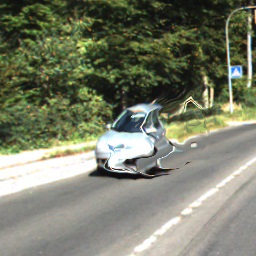"}}
\subfloat[XABA(HT+N)]{\includegraphics[clip, trim=20 20 20 20,width=0.25\linewidth]{"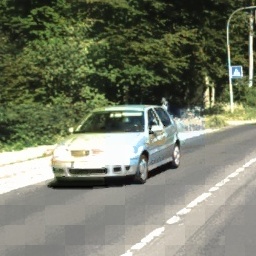"}}
\vspace{-1em}
\end{center}
   \caption{Example Images from Kitti Dataset for Image Alignment. Note that rather than distorting the reference image to match with the target image under planar 1-to-1 constraint, our proposed method prefers transferring a combination of features from the reference to the target image for alignment which better fits the image.}
    \vspace{-1em}
\label{fig:Alignexp}
\end{figure*}

For this experiment, we used images from Kitti dataset paired in time as inputs to XABA and tried to align these images and warp the reference image to the target image. An example pair and alignment results of the different methods can be seen in~\cref{fig:Alignexp}

As shown in~\cref{fig:Alignexp}, optical flow based methods' performance drastically drops whenever there is a large motion. This is basically due to the fact that these methods constraint the change in between frames to planar geometric motion with 1-to-1 pixel correspondence. These constraints from the point of the view of attention mechanism are indeed equivalent to limiting the Attention matrix, $A$, to a permutation matrix where there is one and only one entry being 1 for all of its rows. However, in our case, we can "relax" the permutation matrix constraint by letting the sum of each row to 1 (by using softmax or hard-thresholded ReLU with Normalization), rather than forcing only a single element to be 1 in each row. This relaxation allows contribution of multiple pixels and blending, which warps the reference image to the target image with better performance, which can be seen in \cref{tab:methods}. Effects of different parameters of XABA on image alignment performance can be seen in~\cref{tab:models}. 

\begin{table}[h]
    \centering
    \smallskip
    \begin{tabular}{cc}
        \toprule
        {\textbf{Methods}} & {\textbf{PSNR}}\\
        \midrule
        FlowNet-c~\cite{Dosovitskiy2015} + warp & 18.404 \\
        DISFlow~\cite{Kroeger2016} + warp & 19.891 \\ 
        XABA (Softmax) & \textbf{27.920} \\ 
        \bottomrule
    \end{tabular}
    \caption{PSNR results for different alignment algorithms. Ours has the highest PSNR.}
    \vspace{-1em}
    \label{tab:methods}
\end{table}

\begin{table}[ht]
    \centering
    \smallskip
    \begin{tabular}{ccc}
        \toprule
        \multicolumn{2}{c}{{\textbf{Model parameters}}} & \multirow{3}{*}[0em]{\textbf{PSNR}} \\
        \cmidrule(lr){1-2}
        {\# of interframe} & {Block} & {}\\
        {aligners} & {size} & {}\\
        \midrule
        1 & 10x10 & 20.004\\
        1 &  20x20 & 22.873\\
        \midrule
        2 (1,2) & 10x10 & 23.151\\
        2 (1,2) & 20x20 & 26.090\\
        \midrule
        3 (1,2,4) & 10x10 & 25.149\\
        3 (1,2,4) & 20x20 & \textbf{27.920}\\
        
        \bottomrule
    \end{tabular}
    \caption{PSNR values on Kitti test set for different model parameters. Values in the parenthesis in number of interframe aligners denote the downsample and upsample factors for each interframe aligner. $(f_e,f_m)$ are chosen as (32,16) in interframe aligner block. Non-overlapping block size affects T2B and B2T operations. 3 interframe aligners with block size of 20 perform the best.}
    \label{tab:models}
    \vspace{-1em}
    
\end{table}

\subsection{Super-Resolution Performance}
\label{sec:sr_perf}

To show the effectiveness of XABA as a sub-network of a greater network, we conducted a multi-image super-resolution experiment. Two different images of the same scene is given as an input to a network to find x4 higher resolution image of the same scene. For this experiment, we selected XLSR~\cite{Ayazoglu2021} as the SISR baseline network. Then only by changing the input filters to accept two input RGB images, we constructed so called XLSR\_MISR. By combining XLSR\_MISR with DISFlow+Warp and XABA, we constructed XLSR\_MISR + DIS + Warp and XLSR\_MISR + XABA, respectively. Here for XABA, we used two different activation functions. PSNR values are calculated using the original high resolution images with RGB outputs.

\begin{figure*}
\begin{center}

\subfloat[Original Image]{\includegraphics[clip, trim=100 50 50 100,width=1\linewidth]{"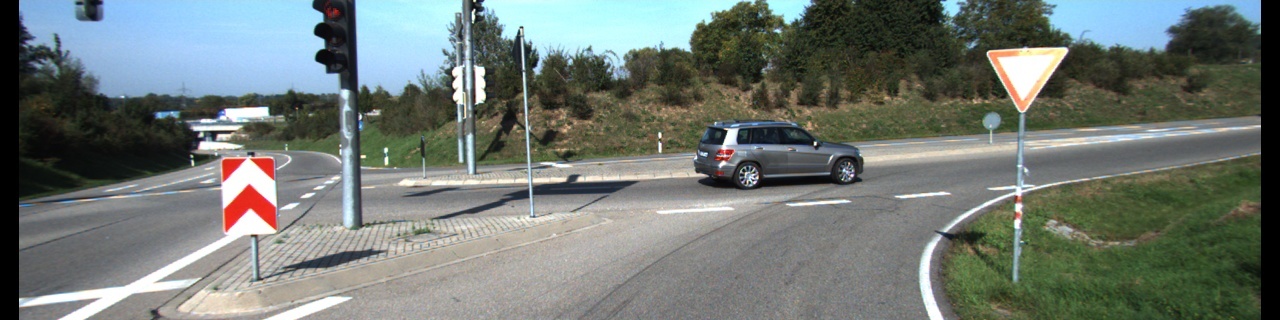"}} \\
\subfloat[Bicubic]{\includegraphics[clip, trim=20 20 20 20, width=0.20\linewidth]{"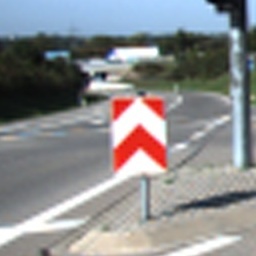"}}
\subfloat[No Alignment]{\includegraphics[clip, trim=20 20 20 20,width=0.2\linewidth]{"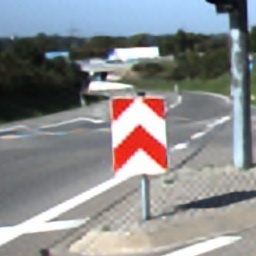"}}
\subfloat[DISFlow]{\includegraphics[clip, trim=20 20 20 20,width=0.2\linewidth]{"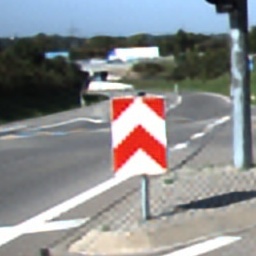"}}
\subfloat[XABA(Softmax)]{\includegraphics[clip, trim=20 20 20 20,width=0.2\linewidth]{"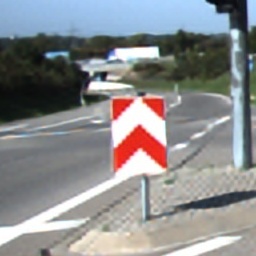"}}
\subfloat[XABA(HT+N)]{\includegraphics[clip, trim=20 20 20 20,width=0.2\linewidth]{"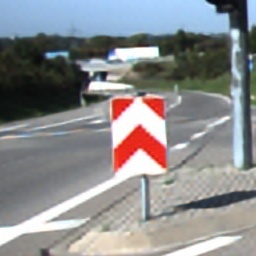"}}

\end{center}
   \caption{Example Images from Kitti Dataset. Note that the lanes are distorted on the other methods while XABA block helps the super-resolution network to better distinguish the lanes, which match with the original image better.}
   
\label{fig:SRexp}
\end{figure*}

As seen in~\cref{tab:sr-psnr}, when an input tensor adjusted SISR method (XLSR\_MISR) is combined with our XABA (HT+N), the performance is improved by 0.1dB. Furthermore, the SISR method is chosen to be with low number of parameters to limit its receptive field to a limited region. This causes degradation on the performance on MISR problem, as shown in XLSR\_MISR without any alignment. The usage of correct alignment with XABA shows the effectiveness of our algorithm. Although DISFlow corrects and aligns the relevant data in the receptive field region, Kitti image pairs usually have large motion and large motions cannot be effectively compensated. An example output of the different alignment methods combined with the super-resolution network can be seen in~\cref{fig:SRexp}.
\begin{table}[h]
    \centering
    \smallskip
    \begin{tabular}{cccc}
        \toprule
        {\textbf{SR Type}} & {\textbf{Alignment}} & {\textbf{PSNR}} & $\Delta${\textbf{ PSNR}}\\
        \midrule
        Bicubic & $\times$ & 23.965 & -1.26\\
        XLSR & $\times$ & 25.223 & 0.00\\
        XLSR\_MISR & $\times$ & 25.216 & -0.01\\
        XLSR\_MISR & DISFlow + Warp & 25.231 & 0.01\\
        XLSR\_MISR & XABA (Softmax) & 25.289 & 0.07\\
        XLSR\_MISR & XABA (HT+N) & \textbf{25.323} & \textbf{0.10}\\
        \bottomrule
    \end{tabular}
    \caption{PSNR values of Kitti test set used in super resolution (SR). In XABA, 3 interframe aligners are used with block size 20x20 and $(f_e,f_m)=(32,16)$. Our method with hard threshold + normalize activation function (HT+N) outperforms the others. HT+N also outperforms Softmax in terms of timing performance for same block sizes (\cref{tab:compGlobalLocalAlign}). }
    \label{tab:sr-psnr}
    \vspace{-2em}
\end{table}

\subsection{Embedded Benchmarks Performance}
\label{sec:embedded_bench_study}
In this experiment, we have investigated the computational performance of our proposed method XABA on an embedded computation device. For this purpose, NVIDIA Jetson AGX Xavier GPU~\cite{NVIDIAJetsonAGX} was used in the benchmark tests. To have a better understanding of the computational performance of the block, the high-performance computing benchmarks are also provided for reference. For the high-performance tests NVIDIA GeForce RTX 3080 GPU~\cite{NVIDIARTX3080} was used. For the inference measurements PyTorch models were exported to ONNX file format, which were then converted to TensorRT engine using NVIDIA TensorRT-Command Line Wrapper, trtexec~\cite{TensorRTExec}. As for TensorRT, we used v7.1.3 and v8.2 for Jetson AGX Xavier GPU and GeForce RTX 3080, respectively. Jetson AGX Xavier GPU has CUDA cores. In this study, we preferred two power consumption modes, which are 15 Watts and 30 Watts for benchmarking on Jetson AGX Xavier GPU. The RTX 3080 GPU has 8704 CUDA cores and consumes 320 Watts as maximum power.


\begin{table*}[ht]
\small
\centering
\begin{tabular}{ccccccccccccc}
\hline
\textbf{Timing Performance$^*$,ms} & \multicolumn{6}{c}{\textbf{Global Alignment Inference}}                                                                                                                                                & \multicolumn{6}{c}{\textbf{Local Alignment Inference}}                                                                                                                                                        \\ \hline
\textbf{Units}                  & \multicolumn{2}{c}{\textbf{AGX (15 W)}}                                 & \multicolumn{2}{c}{\textbf{AGX (30 W)}}                                & \multicolumn{2}{c}{\textbf{RTX 3080}}             & \multicolumn{2}{c}{\textbf{AGX (15 W)}}                                & \multicolumn{2}{c}{\textbf{AGX (30 W)}}                                & \multicolumn{2}{c}{\textbf{RTX 3080}}                     \\ \hline
\textbf{DataType\&Models}       & \multicolumn{1}{c}{\textit{FP16}} & \multicolumn{1}{c}{\textit{FP32}}  & \multicolumn{1}{c}{\textit{FP16}} & \multicolumn{1}{c}{\textit{FP32}} & \multicolumn{1}{c}{\textit{FP16}} & \textit{FP32} & \multicolumn{1}{c}{\textit{FP16}} & \multicolumn{1}{c}{\textit{FP32}} & \multicolumn{1}{c}{\textit{FP16}} & \multicolumn{1}{c}{\textit{FP32}} & \multicolumn{1}{c}{\textit{FP16}} & \textit{FP32}         \\ \hline
XABA20fp32Soft                    & \multicolumn{1}{c}{594.2}         & \multicolumn{1}{c}{904.4}          & \multicolumn{1}{c}{304.8}         & \multicolumn{1}{c}{473.2}         & \multicolumn{1}{c}{39.9}          & 64.5          & \multicolumn{1}{c}{381.4}         & \multicolumn{1}{c}{596.3}         & \multicolumn{1}{c}{202.4}         & \multicolumn{1}{c}{322.7}         & \multicolumn{1}{c}{26.4}          & 42.9                  \\ 
XABA10fp32Soft                           & \multicolumn{1}{c}{368.5}         & \multicolumn{1}{c}{556.9}          & \multicolumn{1}{c}{186.6}         & \multicolumn{1}{c}{282.2}         & \multicolumn{1}{c}{22.1}          & 41.7          & \multicolumn{1}{c}{210.5}         & \multicolumn{1}{c}{339.5}         & \multicolumn{1}{c}{110.9}         & \multicolumn{1}{c}{176.9}         & \multicolumn{1}{c}{14.2}          & 25.4                  \\ 
XABA10fp16Soft                   & \multicolumn{1}{c}{86.7} & \multicolumn{1}{c}{121.9} & \multicolumn{1}{c}{\textbf{43.5}} & \multicolumn{1}{c}{{61.4}} & \multicolumn{1}{c}{{6.5}}  & {10.2} & \multicolumn{1}{c}{46.7} & \multicolumn{1}{c}{{69.7}} & \multicolumn{1}{c}{{24.2}} & \multicolumn{1}{c}{{36.5}} & \multicolumn{1}{c}{{4.4}}  & {6.2}          \\ 
XABA20fp16Soft                   & \multicolumn{1}{c}{146.7}         & \multicolumn{1}{c}{215.1}          & \multicolumn{1}{c}{74.5}          & \multicolumn{1}{c}{111.2}         & \multicolumn{1}{c}{9.9}           & 15.9          & \multicolumn{1}{c}{91.6}          & \multicolumn{1}{c}{140.7}         & \multicolumn{1}{c}{48.1}          & \multicolumn{1}{c}{76.4}          & \multicolumn{1}{c}{7.4}           & 10.5                  \\ 
XABA10fp16HTN              & \multicolumn{1}{c}{{74.1}}              & \multicolumn{1}{c}{{110.5}}               & \multicolumn{1}{c}{\textbf{38.6}}              & \multicolumn{1}{c}{{56.1}}              & \multicolumn{1}{c}{{5.9}}              &      \multicolumn{1}{c}{{9.8}}          & \multicolumn{1}{c}{{40.9}}              & \multicolumn{1}{c}{{64.9}}              & \multicolumn{1}{c}{{20.5}}              & \multicolumn{1}{c}{{32.7}}              & \multicolumn{1}{c}{{4.3}}              & \multicolumn{1}{c}{{5.9}} \\ 
XABA20fp16HTN               & \multicolumn{1}{c}{120.2}              & \multicolumn{1}{c}{193.6}               & \multicolumn{1}{c}{61.1}              & \multicolumn{1}{c}{102.7}              & \multicolumn{1}{c}{8.9}              &        \multicolumn{1}{c}{15.3}       & \multicolumn{1}{c}{72.6}              & \multicolumn{1}{c}{126.9}              & \multicolumn{1}{c}{36.5}              & \multicolumn{1}{c}{65.4}              & \multicolumn{1}{c}{6.2}              & \multicolumn{1}{c}{12.4} \\ \hline
\end{tabular}
     \raggedright \footnotesize{{\bf*}With the contributions of Alperen Kalay (alperenkalay@aselsan.com.tr), Aselsan Research.}
     
    \caption{The Global \& Local Alignment Inference Benchmark Results. The name of the model is encoded as XABA[block\_ size][fp32|fp16][Soft|HTN]. Since the Global Alignment Inference includes Local Alignment Inference computation, we investigate only Global Alignment Inference for real-time performance. The bold results indicate real-time performance.}
    \label{tab:compGlobalLocalAlign}
\end{table*}
The inference benchmark results of image alignment models were separately produced with floating-point16 (FP16) and floating-point32 (FP32) operations. All the training has been conducted with FP32, as it is known that FP16 inference most of the time does not hurt the performance while being two times computationally light~\cite{CudaFP16}. 

\subsection{Timing Results and Power-Efficiency Analysis}
The inference performances of different configurations of the proposed model are comparatively demonstrated in this section. Note that~\cref{tab:compGlobalLocalAlign} indicates timing performances of global and local alignment models. It can be seen that using FP16 for inference is faster than using FP32. On the other hand, the global alignment inference performance can reach about 170 frame per seconds (FPS) on high performance computing device, which is included in the table as a reference to compare it with Jetson Xavier's performance.

As it can be seen from the~\cref{tab:compGlobalLocalAlign}, it is possible to run the proposed block in real-time on Jetson for some specific configuration such as XABA10fp16Soft which indicates the non-overlapping block size for T2B operator is 10 and FP16 is used for inference and activation in Attention module is softmax. Note that by using Hard-Thresholding we could increase the performance of the similar block using softmax in all of the cases and this simple change added almost +3FPS to most of the configurations. Note that local alignment section is also given in the table to get a better understanding of the effects of pyramidal processing and baseline block. Also note that local alignment block can be seen as a pyramid with 0 depth and can still have meaningful usages for aligning smaller displacements such as aligning frames of a video stream.

\begin{figure}[h]
    \centering
    \includegraphics[width=\linewidth]{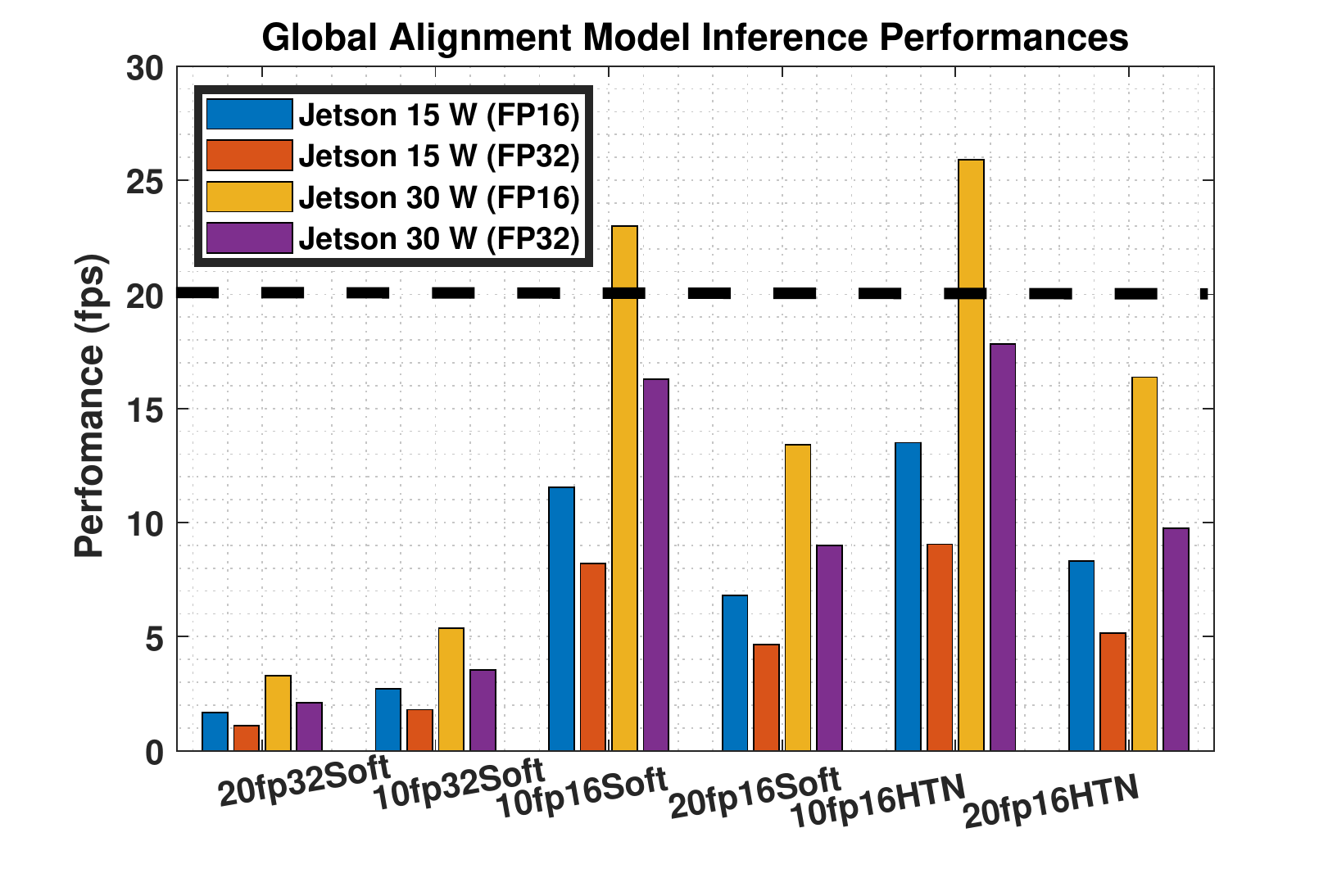}
    \vspace{-2em}
    \caption{The global alignment inference performance benchmarks in terms of FPS for different block configurations.}
    \label{fig:jetsonGlobalAlignFps}
    \vspace{-1em}
\end{figure}

\begin{figure}[h]
    \begin{center}
    \includegraphics[width=\linewidth]{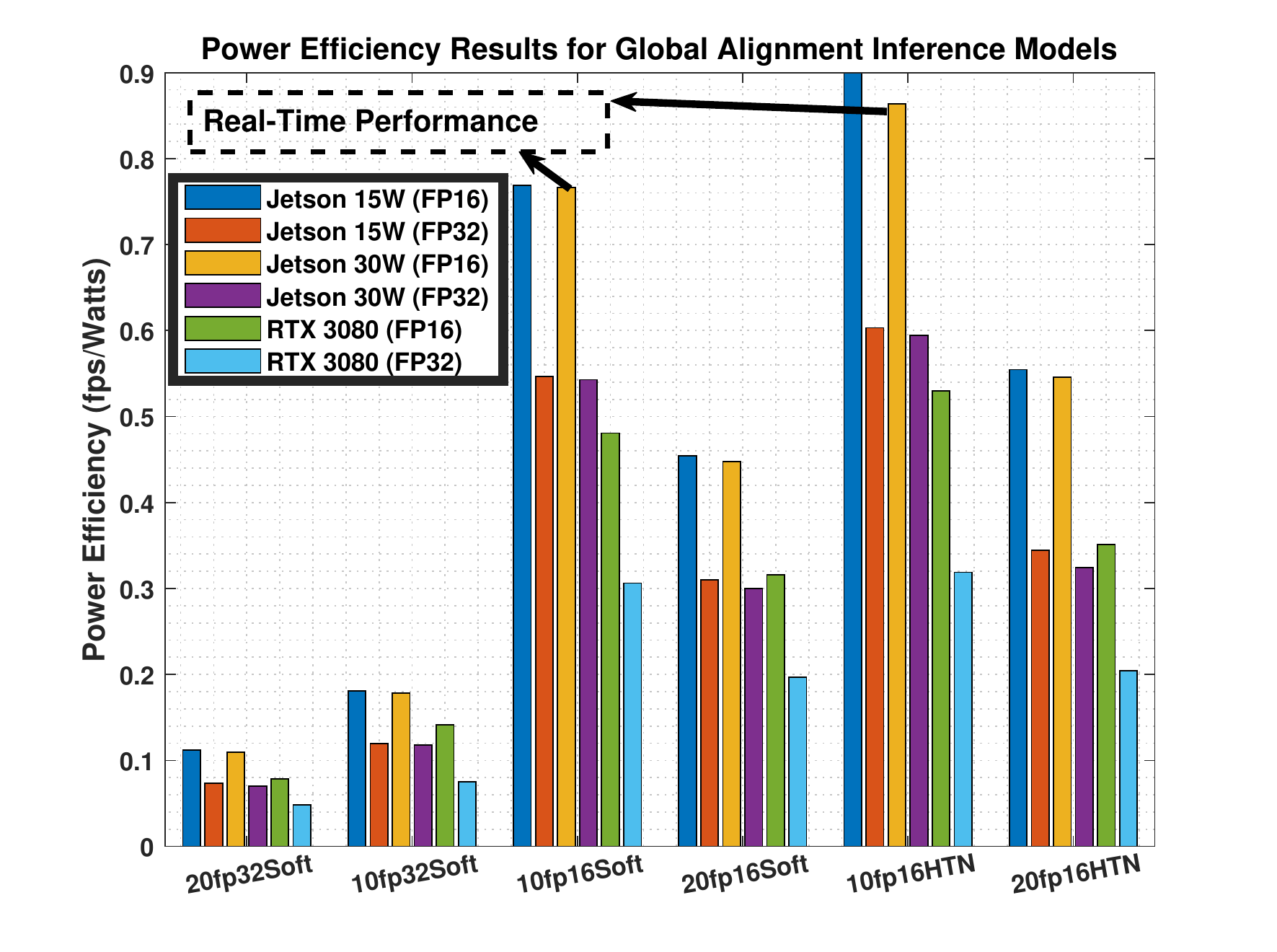}
    \vspace{-2em}
    \caption{The global alignment inference power efficiency for different computation designs.}
    \label{fig:jetsonGlobalAlignPowEff}
    \end{center}
    \vspace{-2em}
\end{figure}

~\cref{fig:jetsonGlobalAlignFps} presents FPS performance results of global alignment inference models, while~\cref{fig:jetsonGlobalAlignPowEff} demonstrates the power efficiency performances for global alignment models.

The computing performance of XABA10fp16Soft inference model meets real-time requirement in terms of FPS for image alignment processing according to the results in~\cref{fig:jetsonGlobalAlignFps}. Indeed, the bottleneck analysis was performed with NVIDIA Profiler for this model. According to bottleneck analysis, the computing time of softmax layer showed that this layer has dominant computation compared to other layers. This inspired our Hard-Thresholding activation proposal which replaces softmax function in activation function. The global alignment inference performance has been increased to 26 FPS from 23 FPS by using hard thresholding and FP16 precision in XABA10fp16HTN. This performance result provides 10\% speedup compared to previous inference model (XABA10fp16Soft).

The power efficiency experiments proved that Jetson AGX GPU provides the most power efficient computation with respect to comparative results in~\cref{fig:jetsonGlobalAlignPowEff}. According to~\cref{fig:jetsonGlobalAlignPowEff}, the Jetson AGX GPU performance provides about 1.6x power efficiency compared to RTX3080 GPU under FP16 precision. From this point of view, this edge device meets real-time image alignment processing requirement in FP16 inference case with 30 Watts power consumption. Although RTX 3080 GPU has high computation performance, its power efficiency performance gave drastically lower result compared to Jetson AGX GPU.

\section{Conclusion}
    
    In this study, we proposed our multi-purpose, cross-attention based image alignment block, XABA. By processing the images in blocks inside a pyramidal block based alignment structure, we capture local relationships with minimal computational need. Focusing on efficiency, we further prove with tests that XABA can run in real-time on edge devices such as NVIDIA Jetson Xavier.
    
    Our experiments reveal that XABA can outperform common optical-flow based alignment methods. We have also shown that XABA can be used as a sub-network aligner in larger deep-learning based scenarios like single and multi-image super-resolution with good performance. Embedded benchmarks and power analyses further prove that pyramidal structure of XABA allows us to realize a power-efficient image aligner.  
    

{\small
\bibliographystyle{ieee_fullname}
\bibliography{egbib}
}

\end{document}